\documentclass{ifacconf}

\usepackage{graphicx}      
\usepackage{natbib}        

\usepackage[utf8]{inputenc}
\usepackage[T1]{fontenc}

\usepackage{algorithm}
\usepackage{algpseudocode}

\usepackage{url}
\usepackage{enumitem}
\usepackage{sidecap}
\usepackage{comment}

\usepackage{graphicx}
\usepackage{dcolumn}
\usepackage{amsmath}
\usepackage{comment} 
\usepackage{amssymb}
\usepackage{mathrsfs} 
\usepackage[normalem]{ulem}
\usepackage{bm}

\usepackage[dvipsnames]{xcolor}

\usepackage{mathtools}
\renewcommand{\doteq}{\vcentcolon=}

\usepackage{cancel}
\usepackage{rotating}
\usepackage{listings}
\usepackage{tagging}
\usepackage{hyphenat}
\usepackage{multirow}

\date{\today}

\begin{document}
	\begin{frontmatter}
		
		\title{Space-Time Diffusion Bridge
		} 
		
		\thanks[footnoteinfo]{We acknowledge use of ACCESS/NSF cloud computing resources via Discover grant {\it Innovation in Generative Diffusion Modeling}, MTH240001.}
		
		\author{Hamidreza Behjoo \& Michael (Misha) Chertkov}

		\address{Program in Applied Mathematics \& Department of Mathematics, University of Arizona, Tucson, AZ 85721, USA \\(email: [hbehjoo,chertkov]@arizona.edu)}

		\begin{abstract}                
			In this study, we introduce a novel method for generating new synthetic samples that are independent and identically distributed (i.i.d.) from high-dimensional real-valued probability distributions, as defined implicitly by a set of Ground Truth (GT) samples. Central to our method is the  integration of space-time mixing strategies that extend across temporal and spatial dimensions. Our methodology is underpinned by three interrelated stochastic processes designed to enable optimal transport from an easily tractable initial probability distribution to the target distribution represented by the GT samples: (a) linear processes incorporating space-time mixing that yield Gaussian conditional probability densities, (b) their diffusion bridge analogs that are conditioned to the initial and final state vectors, and (c) nonlinear stochastic processes refined through score-matching techniques. The crux of our training regime involves fine-tuning the nonlinear model, and potentially the linear models -- to align closely with the GT data. We validate the efficacy of our space-time diffusion approach with numerical experiments, laying the groundwork for more extensive future theory and experiments to fully authenticate the method, particularly providing a more efficient (possibly simulation-free) inference.
		\end{abstract}
		
		
	\end{frontmatter}
	
	\section{Introduction}

	This work addresses how to create new synthetic samples that are indistinguishable from a set of Ground Truth (GT) samples, assumed to be independently and identically distributed (i.i.d.) from an unknown probability distribution. Recently, diffusion models, leveraging Stochastic Differential Equations (SDE) to transform a tractable probability distribution into one that mirrors the GT distribution—thereby embodying the principles of optimal transport—have revolutionized this domain, surpassing competing methodologies \cite{sohl-dickstein_deep_2015,ho_denoising_2020,song_score-based_2021}. Despite their empirical success, a comprehensive theoretical framework for diffusion modeling remains elusive. Our contribution lies in the development of such a framework, enhancing understanding while also introducing innovative learning and inference algorithms.
	
	The essence of the diffusion model lies in its strategy to transform samples over an artificial timeline, adjusting phase-space mixing to achieve both repulsion among distinct samples and convergence towards Ground Truth (GT) samples, often in separate phases of the dynamics. This process, as practiced currently, unfolds in two stages. Initially, we introduce a straightforward stochastic process that indiscriminately mixes all elements of the high-dimensional state vector in a uniform manner, independent of the samples. This process is deemed tractable as its marginal probability density can be computed at any time without the need for simulation. Subsequently, the second stage involves the construction and training of a more intricate stochastic process that not only is influenced by the GT samples but also incorporates both temporal and spatial (inter-component) mixing which is nonlinear in the state variable. The nonlinearity is represented via score function represented via a Deep Neural Network (DNN).

	The core contributions of this manuscript lie in our innovative approach to decouple the traditional sequential relationship inherent in diffusion model construction. We introduce a methodology that embeds space-time mixing directly within the base process from the outset, allowing for subsequent optimization using Ground Truth (GT) samples. A distinctive feature of our work is the parallel exploration of two secondary process options for the second stage: one that progresses forward in time, aligning with the primary process's temporal direction, and another that operates in reverse time. Both secondary options are grounded in the concept of the diffusion bridge, originally introduced by \cite{schrodinger_sur_1932}, related to optimal transport approaches \cite{chen_relation_2014}, and recently utilized in score-based diffusion models \cite{de_bortoli_diffusion_2021, peluchetti_non-denoising_2021, zhou_denoising_2023, li_bbdm_2023}. Bridging from the target distribution, represented by Ground Truth (GT) samples, to a tractable distribution (or vice versa) is advantageous as it eliminates ambiguity in the temporal duration of the episode, thereby fixing its duration.
	
	Let us now highlight this manuscript's contribution to advancing the space-time mixing of images within the forward part of the score-based diffusion process. We will discuss key contributions from others and then differentiate our approach from these notable works. Following the terminology of the discussed papers, we will refer to the spatial and temporal aspects of image mixing as blurring and diffusing, respectively.
	\begin{itemize}
		\item \cite{rissanen2023generative} suggested using the heat equation for blurring (the spatial part of the forward mixing) and then empirically building the reverse process for de-blurring by training the score function (represented via a DNN) to minimize the mismatch between marginal probabilities in the forward and reverse processes. In this approach, noising/de-noising is done separately from blurring/de-blurring as distinct operations. This is fundamentally different from our approach, where blurring and noising are gracefully integrated as part of the space-time mixing of images.
		
		\item The approach of \cite{hoogeboom2023blurring} is built on the premise that blurring can be interpreted as non-isotropic Gaussian diffusion in the Fourier frequency/$k$-space domain. A special $k$-space schedule was developed to keep blurring and noising independent of each other. The transition between pixel space and $k$-space was implemented via the Discrete Cosine Transform (DCT) and its inverse. Notably, our space-time mixing (blurring + noising) is not split into two independent components. Instead, they are correlated in our approach. While aiming to combine computational efficiency, we chose to work in the pixel space in this paper. However, in the $k$-space interpretation, the amount of noise added to each $k$-space component is based on the eigenvalues of the blurring matrix.
		
		\item In the approach closest to ours, \cite{anonymous2022diffusion} combined blurring and noising techniques through stochastic differential equations (SDEs), transforming data into noise using joint corruption and then employing time reversal to generate new samples. Our approach's edge over this prominent technique lies in building the space-time mixing of images within the bridge-representation. Additionally, while \cite{anonymous2022diffusion} utilized the Fourier transform to find the Markov transition kernel, adding an extra layer of complexity, our work deals with the Markov transition kernel directly, avoiding the computationally expensive Fourier transform. Furthermore, we provided thorough experimental validation of our space-time diffusion bridge methodology on the state-of-the-art CIFAR10 dataset, whereas \cite{anonymous2022diffusion} did not include such a comparison.
	\end{itemize}
	We would like to emphasize that the connection between blurring (the spatial part of mixing) and noising (the temporal part of mixing) is a unique contribution of this manuscript—reported in the literature for the first time, to the best of our knowledge.

	We structure the remainder of the manuscript as follows. In Section \ref{sec:affine-DB}, we introduce the foundational concepts of affine (linear in space) space-time diffusion bridges, setting the stage for more complex models and discussions. Section \ref{sec:nonlinear-DB} delves into the nonlinear aspects of space-time diffusion bridges, facilitated by score-matching functions parameterized through DNNs. This section begins with a concise review of relevant literature on score-matching in both diffusion and diffusion bridge processes. We then present two distinct diffusion bridge approaches: a non-denoising scheme operating forward in time in Section \ref{sec:forward} and a denoising scheme in reverse time in Section \ref{sec:reverse}. 
	
	Section \ref{sec:optimal-affine-DB} explores optimizing these diffusion bridge models to enhance alignment with Ground Truth (GT) data. This section is subdivided into an investigation of affine space-time drift optimization in Section \ref{sec:ELBO-optimal-affine} and a heuristic approach for deriving GT-dependent drifts from DNN-parameterized score-functions in Section \ref{sec:optimal-via-score}. Our experimental findings, which validate the theoretical frameworks discussed, are presented in Section \ref{sec:experiments}, covering both MNIST and CIFAR-10 datasets. Concluding remarks and future directions are outlined in Section \ref{sec:conclusions}.
	
	The appendices provide in-depth technical details. Basic derivations of diffusion bridge processes based on Doob's h-transform are presented in Section \ref{app:Bridge-Diffusion}. Discussions on general normal stochastic processes with space-time drift and their diffusion bridge counterparts are detailed in Sections \ref{app:SEN}, \ref{app:SEN-DB}, and \ref{app:direct-NDB}, offering insights into their dual derivations.

	\section{Basic and Diffusion Bridge Processes with Affine Space-Time Mixing Drift}\label{sec:affine-DB}
	
	We examine a class of stochastic Diffusion Bridge (DB) processes for $t\in [0,1]$ where each state vector ${\bm x}(t)$ resides in $\mathbb{R}^k$, with $k$ representing, for example, equal the number of pixels in gray image. The foundational structure of a DB process, elaborated in Appendix \ref{app:Bridge-Diffusion}, originates from a primary stochastic model governed by Eq.~(\ref{eq:SDE}). This model incorporates a deterministic drift vector ${\bm f}(t;{\bm x}(t))\in \mathbb{R}^k$ and a Wiener-diffusion matrix ${\bm \kappa}(t;{\bm x}(t))\in \mathbb{R}^{k\times k}$, which is positive-definite and $k\times k$-dimensional. The process undergoes further refinement by introducing an adjustment to the drift component, ensuring the emergence of a DB process that effectively acts as a conduit between two fixed points—transitioning from an initial state ${\bm x}(0)$ to a terminal state ${\bm x}(1)$, each represented by $\delta$-functions.
	
	In this work, we focus on a particular instance of the basic stochastic process characterized by an affine drift, $\bm{f}(t;\bm{x}(t)) \to \underline{\bm{A}}(t)\bm{x}(t)$, where $\underline{\bm{A}}(t) \in \mathbb{R}^{k \times k}$ denotes a time-dependent, positive-definite matrix independent of $\bm{x}(t)$. Correspondingly, we simplify the Wiener diffusion matrix to $\bm{\kappa}(t; \bm{x}(t)) \to \bm{\kappa}(t) \in \mathbb{R}^{d \times d}$, leading to the following Stochastic Differential Equation (SDE):
	\begin{align} \label{eq:basic-under}
		\forall t \in [0,1]:\quad d\bm{x}(t) &= \underline{\bm{A}}(t)\bm{x}(t)dt + d\bm{W}(t), \\
		\forall i,j:\ \mathbb{E}[dW_i(t) dW_j(t)] &= \kappa_{ij}(t) dt,
	\end{align}
	with $\bm{W}(t)$ representing the Itô-regularized Wiener process, initiating the dynamics at $\bm{x}(0)$. This model is parameterized by $\bm{\psi} \doteq \{\underline{\bm{A}}(0 \to 1), \bm{\kappa}(0 \to 1)\}$. Furthermore, the model may commence from any $\bm{x}(t')$ for $t' \in [0,1]$, proceeding for $t \geq t'$ until $t'' > t'$. While this scenario involves parameters $\{\underline{\bm{A}}(t' \to t''), \bm{\kappa}(t' \to t'')\}$, we consistently employ the notation $\bm{\psi}$ for simplicity, sidestepping additional complexity.
	
	The linearity of SDE as described by Eq.~(\ref{eq:basic-under}) ensures that the solution's statistics, ${\bm x}(t)$, given an initial value ${\bm x}(0)$, adhere to a Gaussian distribution:
	\begin{align} \label{eq:basic-under-PDF}
		&\forall t, t' \in [0,1];\ t \geq t':\quad p_{\bm\psi}\left({\bm x}(t) | {\bm x}(t')\right) \\ \nonumber 
		&\qquad = {\cal N}\left({\bm x}(t) \Bigg| \underline{\bm \Omega}(t;t'){\bm x}(t'); \int\limits_{t'}^t d\tau \underline{\bm \Omega}(t;\tau){\bm \kappa}(\tau)
		\underline{\bm \Omega}(t;\tau)^T\right),
	\end{align}
	where $\underline{\bm \Omega}(t;\tau)$ represents a deterministic matrix process. This process is intricately linked to $\underline{\bm A}(\cdot)$ as defined by the following Ordinary Differential Equations (ODEs):
	\begin{align} \label{eq:under-Omega}
		& 0 \leq \tau \leq t \leq 1:\quad \frac{d}{dt}\underline{\bm \Omega}(t;\tau) = \underline{\bm A}(t)\underline{\bm \Omega}(t;\tau), \\ \nonumber 
		& \frac{d}{d\tau}\underline{\bm \Omega}(t;\tau) = -\underline{\bm \Omega}(t;\tau)\underline{\bm A}(\tau), \quad \underline{\bm \Omega}(\tau;\tau) = {\bm I},
	\end{align}
	with ${\bm I}$ denoting the $d \times d$ identity matrix.

	We establish the following theorem:
	\begin{thm} \label{th:DB-from-basic}
		The DB adaptation of the fundamental stochastic process, as dictated by the SDE (\ref{eq:basic}), is characterized by:
		\begin{gather}\label{eq:basic-bar}
			t \in [0,1]:\quad d{\bm x}(t) = \bar{\bm A}(t)\left({\bm x}(t) - {\bm x}(1)\right)dt + d{\bm W}(t),
		\end{gather}
		where $\bar{\bm A}(\cdot)$ is formulated in terms of $\underline{\bm A}(\cdot)$ via implicit relationships:
		\begin{align} \label{eq:bar-A-via-Omega}
			\bar{\bm A}(t) &\doteq \frac{d}{dt} \log\left(\underline{\bm \Omega}^{-1}(1;t) - {\bm I}\right), \\ \nonumber 
			\frac{d}{dt}\underline{\bm \Omega}^{-1}(1;t) &= \underline{\bm A}(t)\underline{\bm \Omega}^{-1}(1;t), \quad \underline{\bm \Omega}^{-1}(1;1) = {\bm I}.
		\end{align}
	\end{thm}

	\begin{rem}
		The deterministic linear ODE for $\underline{\bm \Omega}^{-1}(1;t)$, as outlined, aligns with the framework established in Eqs.~(\ref{eq:under-Omega}). This can also be perceived through discrete-time relations as $t \to \Delta n/N, n = 0, \ldots, N$, where the evolution of $\underline{\bm \Omega}(1;\Delta n/N)$ and its inverse are defined in the limit $\Delta \to 0$.
	\end{rem}

	\begin{pf}
		The essence of the proof lies in employing Doob's h-transform to evolve a basic SDE (\ref{eq:basic-under}) into a DB SDE and juxtaposing it against the DB formulation (\ref{eq:basic-bar}). Detailed across Appendices \ref{app:Bridge-Diffusion}, \ref{app:SEN}, \ref{app:SEN-DB}, and \ref{app:direct-NDB}, the proof begins with a generalization of the stochastic process governed by Eq.~(\ref{eq:basic}), extending it to both Eqs.~(\ref{eq:basic-under}) and (\ref{eq:basic-bar}). This generalization undergoes the DB transformation procedure, leading to an equation of special form akin to Eq.~(\ref{eq:basic-bar}). Subsequent steps involve the adaptation of parameters to $\underline{\bm A}$ and ${\bm 0}$, culminating in the derivation of Eq.~(\ref{eq:bar-A-via-Omega}) that implicitly connects $\underline{\bm A}$ and $\bar{\bm A}$. The conclusive part of the proof, which reconciles these relations, is elaborated in Appendix \ref{app:direct-NDB}.
	\end{pf}

	\begin{cor}\label{cor:DB-statistics}
		The statistics of the DB process (\ref{eq:basic-bar}) is given by the Gaussian distribution:
		\begin{align}\label{eq:basic-bar-PDF}
			&p_{\bm \psi}\left({\bm x}(t) | {\bm x}(0); {\bm x}(1)\right) = \mathcal{N}\Bigg({\bm x}(t) \Bigg| \bar{\bm \Omega}(t;0){\bm x}(0) + \\ \nonumber 
			&\left({\bm I} - \bar{\bm \Omega}(t;0)\right){\bm x}(1); \int\limits_0^t dt' \bar{\bm \Omega}(t;t'){\bm \kappa}(t')\left(\bar{\bm \Omega}(t;t')\right)^T\Bigg),
		\end{align}
		where the mean and covariance implicitly depend on the parameter set ${\bm \psi} \doteq \{\underline{\bm A}(0 \to 1); {\bm \kappa}(0 \to 1)\}$, as delineated by Eqs.~(\ref{eq:bar-A-via-Omega}).
	\end{cor}
	
	\section{Diffusion Bridge with Nonlinear, Space-Time Mixing Drift}\label{sec:nonlinear-DB}
	
	In this section, we build upon the foundation laid in the preceding discussion, which centered around linear SDEs with affine drift, particularly leveraging the insights from Theorem \ref{th:DB-from-basic}, to extend our modeling framework into the realm of nonlinear SDEs. This extension is pursued through two distinct methodologies, both underpinned by a DNN enforced score-matching strategy. This approach, initially proposed by \cite{hyvarinen_estimation_2005} and further elaborated in \cite{vincent_connection_2011}, has been adapted into discrete-time models in \cite{sohl-dickstein_deep_2015,ho_denoising_2020}, and subsequently evolved into continuous-time diffusion processes in association with SDEs in \cite{song_score-based_2021}. Remarkably, the continuous-time SDE-based algorithms introduced in \cite{song_score-based_2021} align with the earlier theoretical frameworks for denoising reverse time processes, such as those presented in \cite{anderson_reverse-time_1982}. However, our discussion begins by examining the non-denoising forward time approach, thus generalizing the work of \cite{peluchetti_non-denoising_2021}.
	
	\subsection{(Non-Denoising) Forward Time}\label{sec:forward}
	
	The methodology presented in \cite{peluchetti_non-denoising_2021} focuses on the special case of the Brownian bridge. We generalize this framework to the basic SDE (\ref{eq:basic-under}) allowing a Gaussian probability density that conditions on the initial value ${\bm x}(0)$.
	
	In a data-driven scenario, we assume the initial data ${\bm x}(0)$ is independently and identically distributed (i.i.d.) from a sample-tractable distribution $p_0(\cdot)$, such as one with independent components, $p_0({\bm x}(0)) = \prod_{i=1}^k p_{0;i}(x_i)$. Conversely, the final data ${\bm x}(1)$ is derived i.i.d. from the Ground Truth (GT) dataset ${\bm X}^{(S)} = ({\bm x}^{(s)} | s=1,\cdots,S)$.
	
	This approach, underpinned by Corollary \ref{cor:stochastic-bridge} from Appendix \ref{app:Bridge-Diffusion}, employs a revised version of Eq.~(\ref{eq:SDE-pinned}):
	\begin{align}\label{eq:SDE-nonlinear-forward}
		& t \in [0\to 1]:\quad d{\bm x}(t) = \Big(\underline{\bm A}(t){\bm x}(t)+ \\ \nonumber & {\bm \kappa}(t) {\bm s}^{(f)}_{\bm \theta}(t;{\bm x}(t);{\bm x}(1))\Big)dt + d{\bm W}(t).
	\end{align}
	where ${\bm s}^{(f)}_{\bm \theta}(t;{\bm x}(t);{\bm x}(1))$ denotes the vector score function, parameterized by a DNN and parameter set ${\bm \theta}$. The learning of this function is achieved by solving the optimization problem:
	\begin{align}\label{eq:forward-DNN}
		& \min_{\bm\theta} \mathbb{E}_{\text{FT}}\Big[\!\lambda(t)\Big|{\bm s}^{(f)}_{\bm \theta}(t;{\bm x}(t);{\bm x}(1)) - \\ \nonumber &
		\nabla_{{\bm x}(t)} \log\left(p_{\bm \psi}({\bm x}(1) | {\bm x}(t))\right)\Big|^2\!\Big],\ \text{FT}\doteq \Big\{t\sim U([0,1]); \\ \nonumber & {\bm x}(0)\sim p_0(\cdot); {\bm x}(1)\sim {\bm X}^{(S)}; {\bm x}(t)\sim p_{\bm\psi}(\cdot | {\bm x}(0);{\bm x}(1))\Big\},
	\end{align}
	with the expectation taken over the forward trajectory $\text{FT}$, defined by a set of distributions and sampling strategies for ${\bm x}(0)$, ${\bm x}(1)$, and ${\bm x}(t)$. The weight function $\lambda(t)$ serves as a hyperparameter, typically tuned empirically to enhance model performance.
	
	Upon training the score function to optimal parameters ${\bm \theta}^*$, as determined by Eq.~(\ref{eq:forward-DNN}), Eq.~(\ref{eq:SDE-nonlinear-forward}) is then employed for inference, enabling the generation of new samples at $t=1$.
	
	\subsection{(Denoising) Reverse Time}\label{sec:reverse}
	
	In this subsection, we expand upon the denoising reverse time diffusion bridge methodology as outlined in recent works \cite{zhou_denoising_2023,li_bbdm_2023}, focusing on scenarios where the affine-drift matrix $\underline{\bm A}(t)$ simplifies to a scalar coefficient $\alpha(t)$ multiplied by the identity matrix ${\bm I}$. 
	
	We invert the data-driven setting described previously, selecting initial data from the Ground Truth (GT) dataset $({\bm x}(0) \sim {\bm X}^{(S)})$, while the final data ${\bm x}(1)$ is drawn from a tractable distribution $p_0(\cdot)$. Under this configuration, the reverse time adaptations of Eqs.~(\ref{eq:SDE-nonlinear-forward}) and (\ref{eq:reverse-DNN}) are specified as follows, omitting detailed justifications which are thoroughly covered within the generative diffusion literature (e.g., see the review in \cite{yang_diffusion_2022}): 
	\begin{align}\label{eq:SDE-nonlinear-reverse}
		& t \in [1 \to 0]:\ d{\bm x}(t) = \Big(\bar{\bm A}(t)({\bm x}(t) - {\bm x}(1)) \\ \nonumber 
		& \hspace{2.8cm} - {\bm \kappa}(t) {\bm s}^{(r)}_{\bm \phi}(t; {\bm x}(t); {\bm x}(0))\Big)dt + d\tilde{\bm W}(t), \\ 
		\label{eq:reverse-DNN}
		& \min_{\bm \phi} \mathbb{E}_{\text{RT}}\Big[\lambda(t)\Big|{\bm s}^{(r)}_{\bm \phi}(t; {\bm x}(t); {\bm x}(0)) \\ \nonumber
		& - \nabla_{{\bm x}(t)} \log\left(p_{\bm \psi}({\bm x}(t) | {\bm x}(0); {\bm x}(1))\right)\Big|^2\Big], \\ \nonumber 
		& \text{RT} = \Big\{t \sim U([0,1]); {\bm x}(0) \sim {\bm X}^{(S)}; \\ \nonumber 
		& {\bm x}(1) \sim p_0(\cdot); {\bm x}(t) \sim p_{\bm \psi}(\cdot | {\bm x}(0); {\bm x}(1))\Big\},
	\end{align}
	where $\tilde{\bm W}(t)$ denotes a standard Wiener process in reverse time with statistical properties analogous to ${\bm W}(t)$ from the forward time Eqs.~(\ref{eq:basic-under}) and (\ref{eq:basic-bar}). The probability $p_{\bm \psi}({\bm x}(t) | {\bm x}(0); {\bm x}(1))$ leverages the explicit formulation provided by Eq.~(\ref{eq:basic-bar-PDF}), and ${\bm s}^{(r)}_{\bm \phi}(t; {\bm x}(t); {\bm x}(0))$ represents the reverse vector score function, parameterized by a DNN with parameters ${\bm \phi}$.

	\section{Optimal Affine Diffusion Bridge}\label{sec:optimal-affine-DB} 
	
	The generative models discussed in Section \ref{sec:affine-DB}, which incorporate nonlinear space-time drifts learned from GT data via DNN, demonstrate high expressiveness. However, their effectiveness can vary with the GT data and is inherently tied to the underlying basic Gaussian process, as delineated by SDE (\ref{eq:basic-under}) or its diffusion bridge (DB) counterpart (\ref{eq:basic-bar}). In Section \ref{sec:ELBO-optimal-affine}, we examine the flexibility in selecting $\underline{\bm A}(\cdot)$ and/or $\bar{\bm A}(\cdot)$, utilizing the concept of ELBO (Evidence Lower BOund) optimality. Subsequently, Section \ref{sec:optimal-via-score} presents a heuristic approach for deriving a GT-dependent affine drift. This method leverages an iterative process, employing a Taylor expansion of a series of spatially nonlinear score-functions, to enhance model adaptability to GT data.
	
	\subsection{ELBO-Optimal Affine diffusion bridge}\label{sec:ELBO-optimal-affine}
	
	\begin{defn}[max-FT-ELBO]
		The Evidence Lower BOund (ELBO) Objective associated with diffusion bridge (DB) processes, which map initial to final samples, is defined as:
		\begin{align}\label{eq:O}
			{\cal O} & \left(\{\underline{A}(0\to 1)\};   \{{\bm \kappa}(0\to 1)\};p_0(\cdot)\right)=\\ & \nonumber \mathbb{E}_{\text{FT}}\left[\log\left( p_{\bm \psi}\left({\bm x}(t)|{\bm x}(0);{\bm x}(1)\right)\right)\right],
		\end{align}
		with an implicit dependence on ${\bm \psi} = \{\underline{\bm A}(0\to 1), {\bm \kappa}(0\to 1)\}$, as influenced by Eqs.~(\ref{eq:basic-bar-PDF}), (\ref{eq:bar-A-via-Omega}), and (\ref{eq:under-Omega}), as well as the initial distribution $p_0(\cdot)$. The max-ELBO optimization seeks the optimal affine-drift and diffusion matrix values that maximize this sample-dependent objective (\ref{eq:O}):
		\begin{align}\label{eq:max-ELBO}
			& \left(\{\underline{A}^*(0\to 1)\};  \{{\bm \kappa}^*(0\to 1)\};p_0^*(\cdot)\right)\\ \nonumber &\hspace{0.5cm}=\text{argmax}\ {\cal O} \left(\{\underline{A}(0\to 1)\};  \{{\bm \kappa}(0\to 1)\};p_0(\cdot)\right).
		\end{align}
	\end{defn}
	
	\begin{rem}
		Several insights are provided to clarify the significance and nuances of the ELBO objective and its optimal parametrization, along with potential algorithmic implications:
		\begin{enumerate}
			\item Adhering to variational inference terminology \cite{kingma_auto-encoding_2022}, we refer to the objective in Eq.~(\ref{eq:O}) as ELBO to highlight it as the empirical expectation of the lower bound on the finite-time distribution of ${\bm x}(1)$, implicitly represented through the GT sample set ${\bm X}^{(S)}$.
			\item The framework allows for an extension where $\underline{\bm A}(0\to 1)$ may also depend on ${\bm x}(1)$, resulting in $\underline{\bm A}(0\to 1; {\bm x}(1))$, a matrix function parameterizable via a DNN. This generalization can extend to the diffusion matrix as $\{{\bm \kappa}(0\to 1; {\bm x}(1))\}$.
			\item For evaluation efficiency, it might be beneficial to streamline the parameter space in Eq.~(\ref{eq:max-ELBO}). Options include working with time-independent or co-dimensional matrices $\underline{\bm A}(0\to 1)$, restricting the class of initial distributions to those that are tractable, or tailoring the spatial structures of the affine-drift and diffusion matrices to specific applications, such as employing graph-Laplacian structures for image-related tasks.
		\end{enumerate}
	\end{rem}
	
	The following statement follows from Theorem \ref{th:DB-from-basic} by construction:
	\begin{cor}
		For a class of probability distributions which can be represented by the mapping from $t=0$ of a tractable initial distribution, $p_0({\bm x}_0)$ -- via the basic stochastic affine-drift- diffusion process, governed by Eq.~(\ref{eq:basic-bar}) with a particular $\underline{\bm A}(0\to 1)$ and ${\bm \kappa}(0\to 1)$ -- to $p({\bm x}(1))$, the following statements hold:
		\begin{enumerate}
			\item The DB version of the basic mapping procedure -- where the basic process Eq.~(\ref{eq:basic-bar}) is replaced by its DB version (\ref{eq:basic-under}), with $\bar{\bm A}(0\to 1)$ expressed via $\underline{\bm A}(0\to 1)$ according to Eqs.~(\ref{eq:under-Omega},\ref{eq:bar-A-via-Omega}) and with samples from $p({\bm x}(1))$ used as the pinned values at $t=1$ for $S$ samples -- will converge to the basic mapping procedure in the $S\to\infty$ limit.
			\item The max-ELBO optimization (\ref{eq:max-ELBO}) will result in the optimal values $\Big(\{\underline{A}^*(0\to 1)\}$;  $\{{\bm \kappa}^*(0\to 1)\};p_0^*(\cdot)\Big)$ which converges to the true values in the $S\to \infty$ limit. 
		\end{enumerate}
	\end{cor}
	
	We now propose a bold working hypothesis, pending verification through numerical experiments:
	\begin{conj}\label{conj}
		Initiating with samples ${\bm x}(0)$ drawn from $p_0(\cdot)$ and evolving these samples in accordance with the SDE~(\ref{eq:basic-under}), utilizing optimal affine-drift and diffusion matrices, alongside an initial distribution that satisfies the max-ELBO optimization criterion (\ref{eq:max-ELBO}), 
		we hypothesize that the resulting samples ${\bm x}^{(s)}(1)$ will, in the $S \to \infty$ limit, become i.i.d. according to a probability distribution that either represents a general position or belongs to a sufficiently large and practically relevant subclass thereof, as delineated by the ground truth sample set ${\bm X}^{(S)}$.
	\end{conj}
	
	\subsection{Affine Optimality via (Nonlinear) Score-Function}\label{sec:optimal-via-score}
	
	In this subsection, we explore how the use of a DNN-parameterized score function for the reverse process, as delineated in Eq.~(\ref{eq:reverse-DNN}), provides heuristic alternatives to the direct resolution of Eq.~(\ref{eq:max-ELBO}). While an analogous procedure could be applied to the forward process described by Eq.~(\ref{eq:forward-DNN}), we omit it here for brevity.
	
	Suppose the optimal score function ${\bm s}^{(r)}_{{\bm \phi}^*}({\bm x}(t); {\bm x}(0))$ has been identified. In that case, the optimal affine drift can be derived by approximating this score function with the leading term of its Taylor expansion around the deviation of ${\bm x}(t)$ from ${\bm x}(1)$:
	\begin{align}\label{eq:opt-affine}
		& s^{(r)}_{i; {\bm \phi}^*} (t; {\bm x}(t); {\bm x}(0)) \approx \sum_j (x_j(t) - x_j(0)) H^{(r)}_{ij}(t; {\bm x}(0)), \\
		& \nonumber H^{(r)}_{ij}(t; {\bm x}(0)) = \partial_{x_j(t)} s^{(r)}_{j; {\bm \phi}^*}({\bm x}(t); {\bm x}(0))\Big|_{{\bm x}(t)={\bm x}(0)}.
	\end{align}
	
	We can deduce the right-hand side of Eq.~(\ref{eq:opt-affine}) from the optimal score function, represented by a DNN, using differential programming techniques, such as those discussed in \cite{meng2021estimating}. Alternatively, this can be accomplished by addressing a "spatially-smooth" version of Eq.~(\ref{eq:reverse-DNN}) as follows:
	\begin{align}\label{eq:reverse-Hessian-DNN}
		\min_{\bm \theta} & \mathbb{E}_{\text{RT}}\Bigg[\lambda(t)\sum_{i,j}\Big(H^{(r)}_{ij; \bm \theta}(t; {\bm x}(0)) - \\ \nonumber &
		\nabla_{x_i(t)}\nabla_{x_j(t)} \log\left(p_{\bm \psi}({\bm x}(t) | {\bm x}(0); {\bm x}(1))\right)\Big|_{{\bm x}(t)={\bm x}(0)}\Big)^2\Bigg],
	\end{align}
	thereby estimating the sample-averaged Hessian of the log-probability. 
	
	Solving optimization (\ref{eq:reverse-Hessian-DNN}) yields the Hessian, facilitating a simulation-free approach to inference as outlined in the following pseudo-algorithm for generating new samples, ${\bm x}^{(\text{new})}$:
	\begin{enumerate}
		\item Update $\underline{\bm A}(t; {\bm x}(0)) \leftarrow \underline{\bm A}(t) + H^{(r)}_{{\bm \theta}^*}(t; {\bm x}(0))$, consequently updating ${\bm \psi}$ to be dependent on ${\bm x}(0)$.
		\item Sample ${\bm x}(0) \sim {\bm X}^{(S)}$.
		\item Generate ${\bm x}^{(\text{new})} \sim p_{\bm \theta}({\bm x}^{(\text{new})} | {\bm x}(0); {\bm x}(1))$.
	\end{enumerate}
	This process -- iterative learning of the score function with respect to the current ${\bm \psi}$ (now dependent on ${\bm x}(0)$) and subsequent updating of ${\bm \psi}$ -- can be repeated until convergence, or at least as long as the ELBO objective in (\ref{eq:max-ELBO}) shows improvement.

	\section{Numerical Experiments}\label{sec:experiments}
	
	In our investigation, we undertake numerical experiments utilizing both the MNIST and CIFAR-10 datasets, which are established benchmarks in the machine learning domain.
	
	\subsection{MNIST Experiments}
	
	The MNIST dataset, consisting of 70,000 images with dimensions of \(28 \times 28\) pixels, is used as a benchmark for handwritten digit recognition. For the purpose of training our model, we allocate 60,000 images, reserving the remaining 10,000 images to evaluate the model's capability through the Fréchet Inception Distance (FID) metric. This metric assesses the quality and diversity of generated images by comparing their statistical properties in the feature space of a pre-trained DNN against those of real images. Importantly, a lower FID score signifies a closer approximation to the actual data distribution, reflecting the model's enhanced ability to produce images that are perceived as more realistic by human evaluators. This metric is pivotal for our analysis as it provides a robust indicator of image quality that aligns well with human visual judgment.
	
	For our DNN, we opt for the NCSN++ architecture\footnote{\url{https://github.com/yang-song/score_sde_pytorch}}, recognized for its effectiveness in generative modeling. The training is conducted on an NVIDIA A100 GPU, spanning over 100 epochs to ensure sufficient learning and adaptation of the model. We utilize a batch size of 64, employing the Adam optimizer for its efficiency in handling sparse gradients and adaptability in large-scale problems, with a learning rate set at \(10^{-4}\). This configuration is chosen to balance the computational efficiency and the convergence rate, optimizing the model's performance in generating high-quality images.
	
	Our investigations extend to stochastic Gaussian processes governed by basic Stochastic Ordinary Differential Equations (SDE), outlined in Eq.~(\ref{eq:basic-under}). Additionally, we explore their Diffusion Bridge (DB) variants, delineated by Eq.~(\ref{eq:basic-bar}), alongside non-Gaussian stochastic processes operating in both forward and reverse time. These processes are integral to generative models trained on Ground Truth (GT) data using DNNs, as elaborated in Sections \ref{sec:forward} and \ref{sec:reverse}.
	
\begin{figure}[!htbp]
		\centering
		\includegraphics[scale = 0.75]{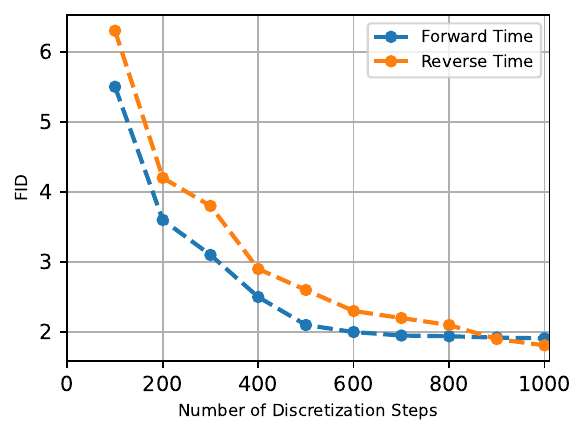}
		\caption{Comparison of FID scores between forward time and reverse time generative models employing the Brownian Bridge scheme, characterized by $\bar{\bm{A}}(t) = \bm{I}/(1-t)$ and $\bm{\kappa}(t) = \bm{I}$. Notably, with 1000 discretization steps, both models achieve an FID score of approximately 2, indicating a high degree of similarity in image quality and diversity to the real data distribution at this level of discretization.}
		\label{fig:bridge-brownian}
	\end{figure}

	We choose the "temporal" model, often referred to as the Brownian Bridge, for the benchmark. This involves DB matrix and covariance matrix which are co-dimensional to the unit matrix: $\bar{\bm{A}}(t) = \bm{I}/(1-t)$ and $\bm{\kappa}(t) = \bm{I}$. As derived from Eqs.~(\ref{eq:basic-bar-PDF}), this model yields a Gaussian process characterized by the mean vector, ${\bm{\mu}}(t) = (1-t) \bm{x}(0) + t \bm{x}(1)$, and the covariance matrix, $\bm{\Sigma}(t) = t(1-t) \bm{I}$. The FID scores for the generative model, tailored to this temporal DB model and applied to MNIST GT data, are depicted in Fig.~\ref{fig:bridge-brownian}. This figure facilitates a comparison between the forward time scheme (discussed in Section \ref{sec:forward}) and the reverse time scheme (outlined in Section \ref{sec:reverse}), revealing that for fewer than 1000 discretization steps, the forward time scheme is more effective. Conversely, the reverse time scheme surpasses the forward scheme in performance when the discretization steps exceed 1000.  
	We adopt $\bar{\bm{A}}(t) = {\bm{L}}/(1-t)$ and $\bm{\kappa}(t) = \bm{I}$ in our space-time mixing model, where ${\bm{L}}$ signifies the graph-Laplacian matrix that models interactions between nearest-neighbor pixels on a square grid. Defined as $\bm{L} = \bm{D} - \bm{J}$, where $\bm{J}$ is the adjacency matrix and $\bm{D}$ a diagonal matrix representing each pixel's connectivity degree, $\bm{L}$ emerges as a symmetric positive semi-definite matrix. Its eigenvalue decomposition $\bm{L} = \bm{P} \text{diag}(\lambda_1, \cdots, \lambda_n) \bm{P}^\top$, where the eigenvalue $\lambda_i \geq 0$. The resultant space-time DB model, as per Eqs.~(\ref{eq:basic-bar-PDF}), produces a Gaussian process with mean vector, $\bm{\mu}(t)= (1-t)^{\bm{L}} \bm{x}(0) + (\bm{I} -(1-t)^{\bm{L}}) \bm{x}(1)$, and covariance matrix, $\bm{\Sigma}(t)=\bm{P} \ \text{diag} \left( \frac{ (1 - t)^{2\lambda_1} - (1 - t) }{1-2\lambda_1 }, \cdots, \frac{ (1 - t)^{2\lambda_n} - (1 - t) }{1-2\lambda_n } \right) \ \bm{P}^T$.
	
\begin{figure}[!htbp]
		\centering
		\includegraphics[scale = 0.75]{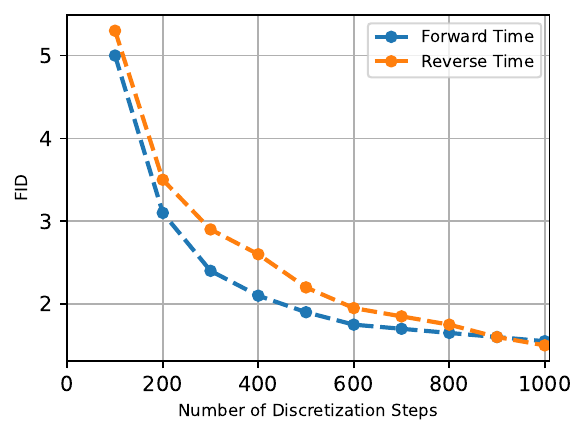}
		\caption{FID scores comparison for space-time DB models employing forward and reverse time approaches. This illustration underscores the models' performance in generative tasks, as elaborated in the accompanying text. Notably, with 1000 discretization steps, the FID score approximates 1.5, highlighting the nuanced efficacy of these models in capturing the data's underlying distribution with a high degree of fidelity.}
		\label{fig:bridge-mnist-fid}
	\end{figure}

\begin{figure}[!htbp]
		\centering
		\includegraphics[scale = 0.5]{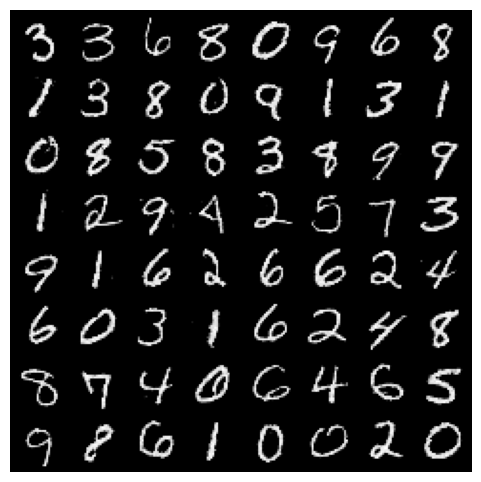}
		\caption{This visualization showcases the model's generative capability to synthesize high-fidelity images by integrating the space-time DB model within the reverse time framework with 1000 discretization steps.}
		\label{fig:bridge-mnist}
	\end{figure}

	In our framework, each image is represented as a grid graph with dimensions $28 \times 28$, leading to a Laplacian matrix of size $784 \times 784$. This approach facilitates the exploration of the space-time DB generative modeling effectiveness, with FID scores presented in Fig.~(\ref{fig:bridge-mnist-fid}). We see that for discretization steps under 1000, the forward time scheme outperforms the reverse time scheme in terms of FID score. Conversely, at 1000 discretization steps, the reverse time scheme demonstrates a marginal advantage, underscoring the nuanced performance dynamics between these schemes. Samples of generated images, with 1000 steps discretizing the $[0,1]$ time interval and using space-time bridge and reverse time, are shown in Fig. \ref{fig:bridge-mnist}.
	
	Comparing Fig.~\ref{fig:bridge-brownian} with Fig.~\ref{fig:bridge-mnist-fid}, we note that space-time mixing achieves superior FID scores using fewer discretization steps. However, this advantage requires evaluating the graph Laplacian, indicating an added computational cost (for training and not inference). To extend this method to higher-resolution images, adopting techniques from auto-encoders, as exemplified in the Stable Diffusion (text-to-image generation) methodology \cite{rombach2021highresolution}, offers a viable pathway for scaling the approach effectively.
	
	\subsection{CIFAR-10 Experiments}
	
	The CIFAR-10 dataset is a widely-used benchmark in the machine learning community for image classification tasks. It consists of 60,000 color images in 10 different classes, with each image having dimensions of \(3\times 32 \times 32\) pixels. The dataset is divided into 50,000 training images and 10,000 test images. Each class contains 6,000 images, providing a diverse and challenging dataset for evaluating the performance of generative models.
	
	We used the same DNN architecture as in \cite{yang_diffusion_2022} to ensure a fair comparison with their results. For this experiment, we focused on the scheme described in Section \ref{sec:reverse}, which includes time reversal in the construction. As observed in the experiments on MNIST reported above, the two schemes—one described in Section \ref{sec:forward} and the other in Section \ref{sec:reverse}—exhibit comparable performance.
	
\begin{figure}[h!]
		\centering
		\begin{tabular}{c}
			\includegraphics[scale=0.75]{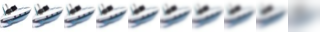} \\
			\includegraphics[scale=0.75]{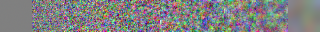} \\
			\includegraphics[scale=0.75]{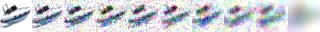} \\    
		\end{tabular}
		\caption{The first row shows the mean of the space-time diffusion bridge, and the second row shows the amount of noise added at each time step. The third column shows the combination of blurring and noise. All results are shown for $t \in [\varepsilon, 1-\varepsilon]$. \label{fig:forward}}
	\end{figure}

	The dynamics of blurring and noising in the forward space-time diffusion bridge process are illustrated in Fig.~\ref{fig:forward}. The figure shows the gradual addition of mean and noise at each step.
	
	\begin{figure}[h!]
		\centering
		\includegraphics[scale=0.85]{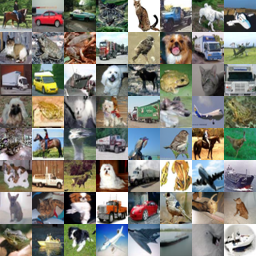}
		\caption{Samples of generated images from the space-time diffusion bridge.}
		\label{fig:cifar-spacetime}
	\end{figure}
	
	\begin{table}[h]
		\centering
		\begin{tabular}{|c|c|c|c|}
			\hline
			& Number of NN Parameters & NFE & FID \\
			\hline
			\cite{yang_diffusion_2022} & 36.9 M & 1000 & 3.77 \\
			\hline
			Ours & 36.9 M & 500 & 2.95 \\
			\hline
		\end{tabular}
		\caption{Comparison of our method with \cite{yang_diffusion_2022}. The number of DNN parameters, number of function evaluations (NFE), and Fréchet Inception Distance (FID) are compared.}
		\label{tab:example}
	\end{table}

	\section{Conclusion and Path Forward}\label{sec:conclusions}
	
	In this work, we introduce a novel space-time generalization of Diffusion Bridge (DB) processes. Our contribution lies in expanding the class of tractable Gaussian processes within the DB framework by incorporating a novel spatial mixing that is linear/affine with respect to the state-vector, in addition to the temporal mixing found in existing literature. We further develop this affine space-time DB concept through the integration of more complex nonlinear dynamics, utilizing score-matching functions and DNNs. Our exploration leads to the proposal of innovative DB schemes, delineating the distinctions between non-denoising (forward in time) and denoising (reverse in time) approaches. Moreover, we investigate the optimization of these schemes to enhance alignment with Ground Truth (GT) data, presenting a theoretical framework for affine space-time drift optimization and the derivation of optimal and GT-dependent affine drifts.
	
	Through empirical validation on the MNIST and CIFAR-10 datasets, we substantiate our theoretical claims. Our comparative analysis reveals that the forward time (without denoising) scheme yields superior results with a sufficiently small number of discretization steps, while both forward and reverse time approaches exhibit comparable Fréchet Inception Distance (FID) performance as the number of steps approaches 1000. The CIFAR-10 experiments, in particular, demonstrate the robustness and scalability of our approach, achieving a lower FID score with fewer DNN function evaluations compared to existing methods. 
	
	Moving forward, our efforts will focus on further experimental validation of the theoretical constructs presented within this manuscript. We aim to delve deeper into the ELBO-based affine drift optimization scheme introduced in Section \ref{sec:ELBO-optimal-affine}, as well as the iterative learning process for the score function, as detailed in Section \ref{sec:optimal-via-score}, applying these methodologies across both forward and reverse time DNN learning approaches for score functions. 
	

	\appendix 
	
	\section{General Diffusion Bridge Processes}\label{app:Bridge-Diffusion}
	
	We commence this Appendix by revisiting the classic problem of the diffusion bridge, closely related to what is known as Doob's h-transform. This topic has been thoroughly explored in the literature, including Chapter 7.5 of \cite{sarkka_applied_2019}, as well as Chapter IV.6.39 in \cite{rogers_diffusions_2000}. And recently in the field of generative AI modeling, as discussed in Section 3.1 of \cite{peluchetti_non-denoising_2021}.
	
	We consider a general drift-diffusion process that governs the stochastic dynamics of a $d$-dimensional vector, denoted by ${\bm x}(t) \doteq  (x_i(t) \in \mathbb{R} \,|\, i=1,\ldots,d)$:
	\begin{align}\label{eq:SDE}
		&t \in [0,1]:\quad d{\bm x}(t) = {\bm f}(t;{\bm x}(t))\,dt + d{\bm W}(t), \\ \nonumber & \forall\, i,j=1,\ldots,d:\ \mathbb{E}[dW_i(t)\,dW_j(t)] = \kappa_{ij}({\bm x}(t),t),
	\end{align}
	where ${\bm W}(t)$ represents a Wiener process characterized by the space-time dependent covariance matrix ${\bm \kappa}({\bm x}(t),t)$. The conditional probability $p_{\bm\psi}({\bm x}(t) | {\bm x}(t'))$, representing the probability of observing ${\bm x}$ at time $t$ given the observation of ${\bm x}'$ at an earlier time $t'$, where $1 > t > t' > 0$, is governed by the Fokker-Planck (FP) equations for forward and reverse dynamics, respectively:
	\begin{align}
		\label{eq:FP-forward}
		& \left(\partial_t - \hat{\cal L}^*_t({\bm x}(t))\right) p_{\bm\psi}({\bm x}(t) | {\bm x}(t')) = 0,\\ \nonumber &
		\hat{\cal L}^*_t({\bm x}) \equiv -\partial_i f_i(t;{\bm x}) + \frac{1}{2}\partial_i \partial_j \kappa_{ij}({\bm x},t),\\ 
		\label{eq:FP-reverse}
		& \left(\partial_t + \hat{\cal L}_t({\bm x}(t))\right) p_{\bm\psi}({\bm x}(1) | {\bm x}(t)) = 0, \\ \nonumber &  
		\hat{\cal L}_t({\bm x}) \equiv f_i(t;{\bm x})\partial_i + \frac{1}{2} \kappa_{ij}({\bm x},t)\partial_i\partial_j,
	\end{align}
	where $\hat{\cal L}$ and $\hat{\cal L}^*$ denote the direct and adjoint generators of the stochastic dynamics, respectively. We adopt the notations and terminology from S\"{a}rkk\"{a} and Solin \cite{sarkka_applied_2019}, utilizing shorthand $\partial_i$ to denote $\partial_{x_i}$ and invoking the Einstein summation convention for repeated indices. The forward \eqref{eq:FP-forward} and reverse \eqref{eq:FP-reverse} forms of the FP equation are well-established within stochastic calculus, most succinctly derived by marginalizing the discretized Feynman-Kac (path-integral) formulations for a joint probability distribution. This approach involves a sequence of Gaussian integrations, followed by taking the continuous-time limit.
	
	Next, we define the transformed probability density
	\begin{equation}\label{eq:p-tilde}
		\tilde{p}({\bm x}(t) | {\bm x}(t')) \doteq  p({\bm x}(t) | {\bm x}(t')) \frac{p({\bm x}(1) | {\bm x}(t))}{p({\bm x}(1) | {\bm x}(t'))},
	\end{equation}
	and note that: (a) it constitutes a legitimate probability density, as it is non-negative and integrates to one over the entire domain of ${\bm x}(t)$; and (b) as $t \to 1$, it converges to the Dirac delta function $\delta({\bm x}(t) - {\bm x}(1))$.

	The remarkable statement is encapsulated in the following theorem:
	\begin{thm}[Fokker-Planck for DB]\label{th:FP-DB}
		The transformed probability density $\tilde{p}({\bm x}(t) | {\bm x}(t'))$, as defined in Eq.~\eqref{eq:p-tilde}, satisfies the diffusion bridge Fokker-Planck equation:
		\begin{align}\label{eq:FP-bridge} &
			\Big(\partial_t - \hat{\cal L}^*_t({\bm x})\\ \nonumber & \hspace{0.5cm} + \partial_{{\bm x}_i(t)} \kappa_{ij}({\bm x}(t), t) s_{j;{\bm x}(1)}({\bm x}(t);t) \Big) \tilde{p}({\bm x}(t) | {\bm x}(t')) = 0,\\ \nonumber & \hspace{0.5cm} s_{j;{\bm x}(1)}({\bm x}(t);t)\doteq\partial_{{\bm x}_j(t)} \log\left(p({\bm x}(1) | {\bm x}(t))\right).
		\end{align}
	\end{thm}
	\begin{pf}
		The proof is algebraic and straightforward. We apply the operator from the left-hand side of Eq.~\eqref{eq:FP-bridge} to the transformed probability density $\tilde{p}({\bm x}(t) | {\bm x}(t'))$, as defined in Eq.~\eqref{eq:p-tilde}. By employing the chain rule for differentiation and subsequently invoking Eqs.~\eqref{eq:FP-forward} and \eqref{eq:FP-reverse}, we represent the temporal derivatives $\partial_t p({\bm x}(t) | {\bm x}(t'))$ and $\partial_t p({\bm x}(1) | {\bm x}(t))$ in terms of the respective spatial derivatives of $p({\bm x}(t) | {\bm x}(t'))$ and $p({\bm x}(1) | {\bm x}(t))$. Upon collecting all terms, they sum to zero, thus completing the proof.
	\end{pf}

	The ensuing corollary is a direct consequence of Theorem~\ref{th:FP-DB} and is also aligned with Theorem 7.11 in S\"{a}rkk\"{a} and Solin \cite{sarkka_applied_2019}:
	\begin{cor}[Stochastic diffusion bridge Process]\label{cor:stochastic-bridge}
		Given the basic SDE as in Eq.~\eqref{eq:SDE}, we now aim to modify the process by conditioning the solution to satisfy ${\bm x}(1)$ at time $t=1$. Additionally, assuming that the conditional probability density $p({\bm x}(1) | {\bm x}(t))$, associated with the basic SDE \eqref{eq:SDE} and defined in Eq.~\eqref{eq:FP-forward}, has been computed as a function of ${\bm x}(t)$. It results in the following stochastic process:
		\begin{align}\label{eq:SDE-pinned}
			& t \in [0,1]:\quad d{\bm x}(t) = \Big({\bm f}(t;{\bm x}(t)) \\ \nonumber & \hspace{0.5cm} + {\bm \kappa}({\bm x}(t), t) \nabla_{{\bm x}(t)} \log\left(p({\bm x}(1) | {\bm x}(t))\right)\Big)dt + d\bm W(t).
		\end{align}
	\end{cor}
	
	\section{Normal Stochastic Process}\label{app:SEN}
	
	Consider the special case of the stochastic drift-diffusion process (\ref{eq:SDE}) with affine drift, ${\bm f}({\bm x}(t);t)= {\bm A}(t){\bm x}(t)+{\bm c}(t)$,  and ${\bm x}(t)$ independent diffusion, ${\bm \kappa}({\bm x}(t);t)\to{\bm \kappa}(t)$. Then evaluation of the basic Stochastic Differential Equation (SDE) -- with the initial condition at $t=0$ pinned/fixed to ${\bm x}(0)={\bm x}_0$:
	\begin{gather}\label{eq:basic}
		t\in[0,1]:\ d {\bm x}(t)\!=\!\left({\bm A}(t){\bm x}(t)\!+\!{\bm c}(t)\right)dt + d{\bm W}(t).
	\end{gather} 
	Statistics of ${\bm x}(t)$, conditioned to the initial value ${\bm x}(0)$ is Gaussian:
	\begin{align} \nonumber
		& p\left({\bm x}(t)|{\bm x}(0)\right)={\cal N}\Bigg({\bm x}(t)\Bigg|{\bm \Omega}(t;0){\bm x}(0) \\ \label{eq:N-ODE-2} &+\int\limits_0^t d t' {\bm \Omega}(t;t'){\bm c}(t');\int\limits_0^t d t' {\bm \Omega}(t;t'){\bm \kappa}(t'){\bm \Omega}(t;t')\Bigg),\\ \label{eq:Omega}
		& t'\in [\tau,t]:\ \frac{d}{dt'}{\bm \Omega}(t';\tau)={\bm A}(t'){\bm \Omega}(t';\tau), \\ \nonumber & \frac{d}{d\tau}{\bm \Omega}(t';\tau)=-{\bm \Omega}(t';\tau){\bm A}(\tau),\quad
		{\bm \Omega}(\tau;\tau)={\bm I},
	\end{align}
	where thus ${\bm \Omega}(t';\tau)$ is a deterministic matrix process which depends implicitly on $({\bm A}(\tau\to t)$; and we shortcut notations omitting dependence of the probability density on ${\bm A}(0\to 1)$ and ${\bm \kappa}(0\to 1)$.

	\vspace{-0.2cm}
	\section{Normal Diffusion Bridge Process}\label{app:SEN-DB}
	
	In this Appendix we combine ideas from Appendix \ref{app:SEN} and Appendix \ref{app:Bridge-Diffusion}. The goal is to construct a Diffusion Bridge process of a special form -- Gaussian process, starting and resulting in $\delta$-function with a specific evolution of the mean vector and of the covariance matrix.  Eq.~(\ref{eq:basic}) with ${\bm A}(t)$ is our starting point here. The conditional probability density of the normal process -- with the initial condition at $\tau$ and the observation point $t$ -- is 
	\begin{align} \nonumber 
		& p^{(\text{SEN})}_{\bm \psi}\!\!\left({\bm x}(t)|{\bm x}(\tau)\right)\!=\!{\cal N}\Big({\bm x}(t)\Big|{\bm \Omega}(t;\tau){\bm x}(\tau)\!+\!\int\limits_\tau^t\!\! d t' {\bm \Omega}(t;t'){\bm c}(t');\\ \label{eq:Green-SEN} & {\bm \Sigma}(\tau)\Big),\quad {\bm \Sigma}(\tau)\doteq 
		\int\limits_\tau^t d t' {\bm \Omega}(t;t'){\bm \kappa}(t')\left({\bm \Omega}(t;t')\right)^T,
	\end{align}
	where ${\bm \Sigma}(\tau)$ is the covariance matrix; and we shortcut notations omitting dependencies on  ${\bm A}(\tau\to t;\tau)$; ${\bm c}(\tau\to t)$; and ${\bm \kappa}(\tau\to t)$. 
	
	Next, following the general construction of Appendix \ref{app:Bridge-Diffusion} we arrive at the diffusion bridge (DB) version of the process described by Eq.~(\ref{eq:Green-SEN}) with the DB pinned at $\tau=t$ to ${\bm x}(t)$:
	\begin{align}\label{eq:DB-SEN}
		& d {\bm x}(\tau)  =\Big(\!{\bm A}(\tau){\bm x}(\tau)+ {\bm c}(\tau)\\ \nonumber & +{\bm s}\left({\bm x}(\tau);\tau;{\bm x}(t);\{{\bm \theta}(\tau\to t)\}
		\right)\Big) d\tau\! +d{\bm W}(\tau)\\ \label{eq:psi-DB-SEN} 
		& {\bm s} \left({\bm x}(\tau);\tau;{\bm x}(t);\{{\bm \theta}(\tau\to t)\}\right) \\ \nonumber & \doteq 
		\nabla_{{\bm x}(\tau)} \log\left(
		p^{(\text{SEN})}\left({\bm x}(t)|{\bm x}(\tau);\{{\bm \theta}(\tau\to t)\}\right)\right) \\ \nonumber & = - \left({\bm \Sigma}(\tau) \right)^{-1}\left({\bm x}(t)-{\bm \Omega}(t;\tau){\bm x}(\tau)-\int\limits_{\tau}^t d t' {\bm \Omega}(t;t'){\bm c}(t')\right).
	\end{align} 
	We observe that the DB-correction, that is the ${\bm s}$-term, results in ``re-normalization" of the ${\bm A}(\cdot)$ and ${\bm c}(\cdot)$  (functions-) coefficients and can thus be viewed as a correction which allows to restate the DB-SEN Eq.~(\ref{eq:DB-SEN}) as a ``renormalization" of the standard SEN process described by Eq.~(\ref{eq:basic}):
	\begin{align} \nonumber 
		d {\bm x}(\tau) & \!=\! \Big(\tilde{\bm A}(\tau) {\bm x}(\tau)\!-\!\left({\bm \Sigma}(\tau) \right)^{-1}{\bm x}(t)\!+\!{\bm \varsigma}(\tau)\Big)d\tau +d {\bm W}(\tau),\\ \nonumber 
		\tilde{\bm A}(\tau) & \doteq {\bm A}(\tau)+\left({\bm \Sigma}(\tau) \right)^{-1}{\bm \Omega}(t;\tau),\\ {\bm \varsigma}(\tau) & \doteq {\bm c}(\tau)+\left({\bm \Sigma}(\tau) \right)^{-1}\int\limits_{\tau}^t d t' {\bm \Omega}(t;t'){\bm c}(t'). \label{eq:DB-SEN-as-SEN}
	\end{align}
	Then, the conditional probability density associated with Eq.~(\ref{eq:DB-SEN}) 
	becomes:
	\begin{align}\label{eq:GF-SEN-DB}
		& p^{(\text{SEN-DB})} \left({\bm x}(\tau)|{\bm x}(0);{\bm x}(t)\right)={\cal N}\Big({\bm x}(\tau)\Big|\tilde{\bm \Omega}(\tau;0){\bm x}(0) \\  \nonumber & - 
		\tilde{\bm \Lambda}(\tau){\bm x}(t) +\tilde{\bm \varsigma}(\tau);
		\tilde{\bm \Sigma}(\tau)\Big.\Big),\ 
		\tilde{\bm \Lambda}(\tau) \! \doteq \!\!
		\int\limits_0^\tau \!\! d t' \tilde{\bm \Omega}(\tau;t')
		\left({\bm \Sigma}(t') \right)^{-1},\\ \nonumber 
		& \tilde{\bm \varsigma}(\tau) \!\doteq\! \!\!
		\int\limits_0^\tau \!\! d t' \tilde{\bm \Omega}(\tau;t'){\bm \varsigma}(t'),\ 
		\tilde{\bm \Sigma}(\tau) \!\doteq\! \!\!\int\limits_0^\tau \!\! d t' \tilde{\bm \Omega}(\tau;t'){\bm \kappa}(t')\tilde{\bm \Omega}(\tau;t'),
	\end{align}
	where $\tilde{\bm\Omega}(t';\tau)$ is defined implicitly as a solution of the following matrix differential equation:
	\begin{gather}\label{eq:tilde-Omega}
		\frac{d}{d t'}\tilde{\bm \Omega}(t';\tau)=\tilde{\bm A}(t')\tilde{\bm \Omega}(t';\tau), \quad \tilde{\bm \Omega}(\tau;\tau)={\bm I}.
	\end{gather}
	
	Derivations similar to ones presented in this Appendix were reported earlier in \cite{barczy_representations_2010}.
	
	\section{Normal diffusion bridge Process: Alternative Derivation} \label{app:direct-NDB}
	
	Let us consider a normal stochastic process described by Eq.~(\ref{eq:basic-bar}).
	Then from Eq.~(\ref{eq:basic}), with ${\bm A}(t)$ and ${\bm c}(t)$ substituted by $\bar{\bm A}(t)$ and $-\bar{\bm A}(t){\bm x}(1)$ respectively, and then using the identity, $\int_0^t dt' \bar{\bm \Omega}(t;t')\bar{\bm A}(t')=\bar{\bm \Omega}(t;0)-{\bm I}$, where $\bar{\bm \Omega}(\cdot;\cdot)$ satisfies 
	\begin{align}\label{eq:bar-Omega}
		\forall t\geq\tau,\ & t,\tau \in [0,1]:\ \frac{d}{dt}\bar{\bm \Omega}(t;\tau)=\bar{\bm A}(t)\bar{\bm \Omega}(t;\tau),\\ \nonumber & \frac{d}{d\tau}\bar{\bm \Omega}(t;\tau)=-\bar{\bm \Omega}(t;\tau)\bar{\bm A}(\tau),\quad \bar{\bm \Omega}(\tau;\tau)={\bm I},
	\end{align}
	we arrive at the marginal probability distribution of ${\bm x}(t)$ conditioned to ${\bm x}(0)$ at $t=0$ and to ${\bm x}(1)$ at $t=1$ stated in Eq.~(\ref{eq:basic-bar-PDF}).
	
	We observe that if $\bar{\bm A}(t)$ is such that 
	\begin{gather}\label{eq:basic-bar-cond}
		\forall t\in[0,1]: |\bar{\bm A}(t)|<+\infty,\ \& \  \bar{\bm \Omega}(1;t)\to 0\ \text{at}\ t\to 1,
	\end{gather}
	it makes the process described by Eq.~(\ref{eq:basic-bar}) a Normal Diffusion Bridge (NDB) process. 
	
	Notice that constructing the NDB process described by Eqs.~(\ref{eq:basic-bar-PDF},\ref{eq:basic-bar-cond}) we acted directly -- bypassing the procedure described in  Appendix \ref{app:SEN-DB}, where we first introduced a Normal Diffusion (ND) process and then applied to it the DB construction of Appendix \ref{app:Bridge-Diffusion}. It is thus natural to look for expressing $\bar{A}(t)$ in Eq.~(\ref{eq:basic-bar}) via parameters of the ND process (\ref{eq:basic-under}). Juxtaposing Eq.~(\ref{eq:basic-bar}) term-by-term to Eq.~(\ref{eq:DB-SEN-as-SEN}) with ${\bm A}(\cdot), {\bm \Sigma}(\cdot)$ and ${\bm \Omega}(\cdot;\cdot)$ substituted respectively by $\underline{\bm A}(\cdot), \underline{\bm \Sigma}(\cdot)$ and $\underline{\bm \Omega}(\cdot;\cdot)$, we first of all observe that ${\bm c}(\cdot)$ can be set to ${\bm 0}$, and then arrive at the relations:
	\begin{gather}\label{eq:under-A-bar-A}
		\underline{\bm A}(t)+\left(\underline{\bm\Sigma}(t)\right)^{-1}\underline{\bm \Omega}(1;t)=\bar{\bm A}(t),\ \left(\underline{\bm \Sigma}(t)\right)^{-1}=\bar{\bm A}(t).
	\end{gather}
	Resolving the relations, while also accounting for the "underline" version of Eq.~(\ref{eq:Omega}) counted backwards in time we derive Eq.~(\ref{eq:bar-A-via-Omega}).


\begin{thebibliography}{0}
\providecommand{\natexlab}[1]{#1}
\providecommand{\url}[1]{\texttt{#1}}
\providecommand{\urlprefix}{URL }
\expandafter\ifx\csname urlstyle\endcsname\relax
  \providecommand{\doi}[1]{doi:\discretionary{}{}{}#1}\else
  \providecommand{\doi}{doi:\discretionary{}{}{}\begingroup
  \urlstyle{rm}\Url}\fi

\end{thebibliography}


\begin{thebibliography}{22}
		\providecommand{\natexlab}[1]{#1}
		\providecommand{\url}[1]{\texttt{#1}}
		\providecommand{\urlprefix}{URL }
		\expandafter\ifx\csname urlstyle\endcsname\relax
		\providecommand{\doi}[1]{doi:\discretionary{}{}{}#1}\else
		\providecommand{\doi}{doi:\discretionary{}{}{}\begingroup
			\urlstyle{rm}\Url}\fi
		
		\bibitem[{Anderson(1982)}]{anderson_reverse-time_1982}
		Anderson, B.D. (1982).
		\newblock Reverse-time diffusion equation models.
		\newblock \emph{Stochastic Processes and their Applications}, 12(3), 313--326.
		\newblock \doi{10.1016/0304-4149(82)90051-5}.
		\newblock
		
		\bibitem[{Anonymous(2022)}]{anonymous2022diffusion}
		Anonymous (2022).
		\newblock Diffusion models in space and time via the discretized heat equation.
		\newblock In \emph{Submitted to ICLR Workshop on Deep Generative Models for
			Highly Structured Data}.
		\newblock Under review.
		
		\bibitem[{Barczy and Kern(2010)}]{barczy_representations_2010}
		Barczy, M. and Kern, P. (2010).
		\newblock Representations of multidimensional linear process bridges.
		\newblock arXiv:1011.0067.
		
		\bibitem[{Chen et~al.(2014)Chen, Georgiou, and Pavon}]{chen_relation_2014}
		Chen, Y., Georgiou, T., and Pavon, M. (2014).
		\newblock On the relation between optimal transport and
		Schr{\"o}dinger bridges: {A} stochastic control viewpoint.
		\newblock ArXiv:1412.4430.
		
		\bibitem[{De~Bortoli et~al.(2021)De~Bortoli, Thornton, Heng, and
			Doucet}]{de_bortoli_diffusion_2021}
		De~Bortoli, V., Thornton, J., Heng, J., and Doucet, A. (2021).
		\newblock Diffusion {Schrödinger} {Bridge} with {Applications} to
		{Score}-{Based} {Generative} {Modeling}.
		\newblock \emph{Advances in {Neural} {Information} {Processing}
			{Systems}}, volume~34, 17695--17709.
		\newblock
		
		\bibitem[{Ho et~al.(2020)Ho, Jain, and Abbeel}]{ho_denoising_2020}
		Ho, J., Jain, A., and Abbeel, P. (2020).
		\newblock Denoising {Diffusion} {Probabilistic} {Models}.
		\newblock arXiv:2006.11239.
		
		\bibitem[{Hoogeboom and Salimans(2023)}]{hoogeboom2023blurring}
		Hoogeboom, E. and Salimans, T. (2023).
		\newblock Blurring diffusion models.
		\newblock In \emph{The Eleventh International Conference on Learning
			Representations}.
		
		\bibitem[{Hyvärinen(2005)}]{hyvarinen_estimation_2005}
		Hyvärinen, A. (2005).
		\newblock Estimation of {Non}-{Normalized} {Statistical} {Models} by {Score}
		{Matching}.
		\newblock \emph{Journal of Machine Learning Research}, 6(24), 695--709.
		
		\bibitem[{Kingma and Welling(2022)}]{kingma_auto-encoding_2022}
		Kingma, D.P. and Welling, M. (2022).
		\newblock Auto-{Encoding} {Variational} {Bayes}.
		\newblock arXiv:1312.6114.
		
		\bibitem[{Li et~al.(2023)Li, Xue, Liu, and Lai}]{li_bbdm_2023}
		Li, B., Xue, K., Liu, B., and Lai, Y.K. (2023).
		\newblock {BBDM}: {Image}-to-image {Translation} with {Brownian} {Bridge}
		{Diffusion} {Models}.
		\newblock arXiv:2205.07680.
		
		\bibitem[{Meng et~al.(2021)Meng, Song, Li, and Ermon}]{meng2021estimating}
		Meng, C., Song, Y., Li, W., and Ermon, S. (2021).
		\newblock Estimating high order gradients of the data distribution by
		denoising.
		\newblock 
		\emph{Advances in Neural Information Processing Systems}.
		
		\bibitem[{Peluchetti(2021)}]{peluchetti_non-denoising_2021}
		Peluchetti, S. (2021).
		\newblock Non-{Denoising} {Forward}-{Time} {Diffusions}.
		
		\bibitem[{Rissanen et~al.(2023)Rissanen, Heinonen, and
			Solin}]{rissanen2023generative}
		Rissanen, S., Heinonen, M., and Solin, A. (2023).
		\newblock Generative modelling with inverse heat dissipation.
		\newblock In \emph{The Eleventh International Conference on Learning
			Representations}.
		
		\bibitem[{Rogers and Williams(2000)}]{rogers_diffusions_2000}
		Rogers, L.C.G. and Williams, D. (2000).
		\newblock \emph{Diffusions, {Markov} {Processes} and {Martingales}}.
		\newblock Cambridge University Press, 2 Edition.
		
		\bibitem[{Rombach et~al.(2021)Rombach, Blattmann, Lorenz, Esser, and
			Ommer}]{rombach2021highresolution}
		Rombach, R., Blattmann, A., Lorenz, D., Esser, P., and Ommer, B. (2021).
		\newblock High-resolution image synthesis with latent diffusion models.
		
		\bibitem[{Schrödinger(1932)}]{schrodinger_sur_1932}
		Schrödinger, E. (1932).
		\newblock Sur la théorie relativiste de l'électron et l'interprétation de la
		mécanique quantique.
		\newblock \emph{Annales de l'institut Henri Poincaré}, 2(4), 269--310.
		
		\bibitem[{Sohl-Dickstein et~al.(2015)Sohl-Dickstein, Weiss, Maheswaranathan,
			and Ganguli}]{sohl-dickstein_deep_2015}
		Sohl-Dickstein, J., Weiss, E.A., Maheswaranathan, N., and Ganguli, S. (2015).
		\newblock Deep {Unsupervised} {Learning} using {Nonequilibrium}
		{Thermodynamics}.
		\newblock arXiv:1503.03585.
		
		\bibitem[{Song et~al.(2021)Song, Sohl-Dickstein, Kingma, Kumar, Ermon, and
			Poole}]{song_score-based_2021}
		Song, Y., Sohl-Dickstein, J., Kingma, D.P., Kumar, A., Ermon, S., and Poole, B.
		(2021).
		\newblock Score-{Based} {Generative} {Modeling} through {Stochastic}
		{Differential} {Equations}.
		\newblock arXiv:2011.13456.
		
		\bibitem[{Särkkä and Solin(2019)}]{sarkka_applied_2019}
		Särkkä, S. and Solin, A. (2019).
		\newblock \emph{Applied {Stochastic} {Differential} {Equations}}.
		\newblock Cambridge University Press, 1 Edition.
		\newblock
		
		\bibitem[{Vincent(2011)}]{vincent_connection_2011}
		Vincent, P. (2011).
		\newblock A {Connection} {Between} {Score} {Matching} and {Denoising}
		{Autoencoders}.
		\newblock \emph{Neural Computation}, 23(7), 1661--1674.
		\newblock
		
		\bibitem[{Yang et~al.(2022)Yang, Zhang, Song, Hong, Xu, Zhao, Shao, Zhang, Cui,
			and Yang}]{yang_diffusion_2022}
		Yang, L., Zhang, Z., Song, Y., Hong, S., Xu, R., Zhao, Y., Shao, Y., Zhang, W.,
		Cui, B., and Yang, M.H. (2022).
		\newblock Diffusion {Models}: {A} {Comprehensive} {Survey} of {Methods} and
		{Applications}.
		\newblock arXiv:2209.00796.
		
		\bibitem[{Zhou et~al.(2023)Zhou, Lou, Khanna, and Ermon}]{zhou_denoising_2023}
		Zhou, L., Lou, A., Khanna, S., and Ermon, S. (2023).
		\newblock Denoising {Diffusion} {Bridge} {Models}.
		\newblock arXiv:2309.16948.
		
	\end{thebibliography}
\end{document}